\documentclass{article}

\usepackage[numbers]{natbib}
\usepackage[preprint]{neurips_2020}

\usepackage[utf8]{inputenc} 
\usepackage[T1]{fontenc}    
\usepackage{hyperref}       
\usepackage{url}            
\usepackage{booktabs}       
\usepackage{amsfonts}       
\usepackage{nicefrac}       
\usepackage{microtype}      
\usepackage{graphicx}

\usepackage{wrapfig}
\usepackage{multirow}
\usepackage{xcolor}
\usepackage{appendix}
\usepackage{caption}

\usepackage{xcolor}
\usepackage{amsmath}
\usepackage{accents}
\newlength{\dhatheight}

\usepackage{todonotes}

\newcommand{\ignore}[1]{}
\title{STint: Self-supervised Temporal Interpolation for Geospatial Data}

\author{
  Nidhin Harilal\thanks{Department of Computer Science,
  University of Colorado Boulder, CO, USA}\\
  \And
  Bri-Mathias Hodge\thanks{Department of Electrical, Computer and Energy Engineering, University of Colorado Boulder  \& National Renewable Energy Laboratory (NREL), Boulder, CO, USA
}\\
      \And
    Aneesh Subramanian\thanks{Department of Atmospheric and Oceanic Sciences, University of Colorado Boulder, CO, USA}
    \And
    Claire Monteleoni\thanks{INRIA Paris, France}\;\,\footnotemark[1]
}

\begin{document}
\maketitle

\begin{abstract}
Supervised and unsupervised techniques have demonstrated the potential for temporal interpolation of video data. Nevertheless, most prevailing temporal interpolation techniques hinge on optical flow, which encodes the motion of pixels between video frames. On the other hand, geospatial data exhibits lower temporal resolution while encompassing a spectrum of movements and deformations that challenge several assumptions inherent to optical flow. In this work, we propose an unsupervised temporal interpolation technique, which does not rely on ground truth data or require any motion information like optical flow, thus offering a promising alternative for better generalization across geospatial domains. Specifically, we introduce a self-supervised technique of dual cycle consistency. Our proposed technique incorporates multiple cycle consistency losses, which result from interpolating two frames between consecutive input frames through a series of stages. This dual cycle consistent constraint causes the model to produce intermediate frames in a self-supervised manner. To the best of our knowledge, this is the first attempt at unsupervised temporal interpolation without the explicit use of optical flow. Our experimental evaluations across diverse geospatial datasets show that STint significantly outperforms existing state-of-the-art methods for unsupervised temporal interpolation. 
\end{abstract}

\section{Introduction}
Video interpolation is a technique used to generate intermediate frames between consecutive frames in a video sequence. The primary goal is to enhance the temporal resolution, reduce motion blur, and create smoother transitions between frames. Video interpolation is critical in various fields, including computer vision, video processing, climate modeling, satellite imagery, radar data processing, etc. In the field of computer vision, video interpolation has been widely utilized for frame-rate up-conversion~\cite{choi2000new, kaviani2015frame}, slow-motion video generation~\cite{slomo}, and video stabilization~\cite{choi2020deep}. The applicability of video interpolation extends to climate modeling, where it aids in generating high-resolution temporal data from low-resolution climate models~\cite{lee2014nonparametric, dibike2006temporal}, also referred to as temporal interpolation. Interpolation techniques also prove valuable in analyzing other scientific datasets, like satellite imagery analysis, to fill in missing temporal data and improve frame quality~\cite{vandal2021temporal}. Similarly, it is essential for radar data processing~\cite{nielsen2014numerical}, where it is employed to enhance the resolution and accuracy of the retrieved information.
\begin{figure}[t]
    \centering
    \includegraphics[width=0.7\linewidth]{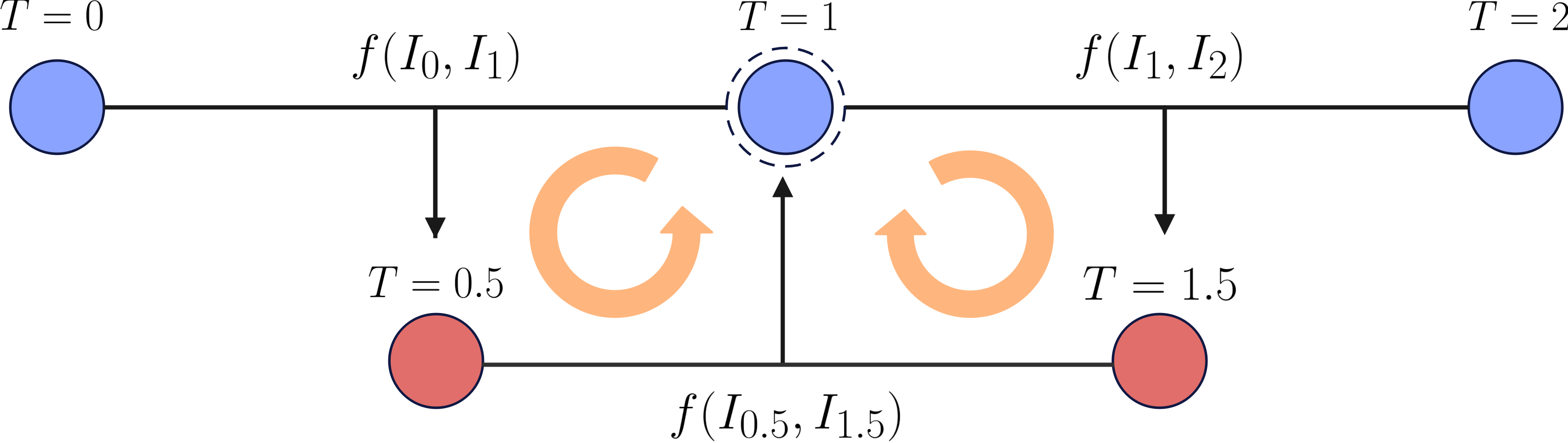}
     \caption{Schematic of cycle consistency technique for temporal interpolation. Here, $I_T$ refers to the $T^{th}$ frame in a temporal sequence. The idea is to use the intermediate frames from a triplet of input sequence to reconstruct back the middle frame. This enforces similarity between frames at $T=1$, which is then used to train a network $f$, imposing a cycle consistency as shown by the two cyclic arrows.}
    \label{fig:cc}
\end{figure}
Most existing methods for video interpolation techniques are supervised and rely on ground truth data to learn the interpolation task. Although a few unsupervised interpolation methods~\cite{unsupcycle, deepcycle} have been proposed in recent years, these leverage optical flow estimation~\cite{opticalflow} to model the motion between consecutive frames, thus limiting them with the assumptions made by optical flow algorithms. Optical flow estimation is sensitive to variations in texture, lighting conditions, and occlusions~\cite{sun2010secrets}. In cases where the motion is non-rigid or exhibits irregular patterns, such as in geospatial datasets, traditional optical flow-based methods may not produce satisfactory results. Therefore, such methods that rely on optical flow can struggle to generalize to other domains, particularly those with complex or unfamiliar motion patterns.

\subsection{Unsupervised Temporal Interpolation}
The need for unsupervised interpolation methods stems from the challenges in addressing the gaps and inconsistencies within datasets, particularly in domains with vast amounts of data, like historical archives of diverse geospatial datasets. In situations where supervised methods are hindered by the lack of labeled data points or where manual interpolation becomes unfeasible due to the sheer volume of data, unsupervised techniques can step in to bridge these gaps effectively. In the context of data-rich scenarios, where historical data might be fragmented or irregularly spaced, unsupervised temporal interpolation methods offer a means to create more continuous and coherent representations of the data. These pre-trained networks can then be fine-tuned in target domains with data-poor situations like inferring higher-temporal information in scarcely recorded regions. By temporally interpolating patterns and relationships within the existing data points, these methods can generate valuable insights into historical variations in geospatial data, aiding in understanding past trends while also extending these insights and predictions to data-poor regions. 

\subsection{Optical Flow Estimation for Geospatial Data}
Geospatial data is inherently complex and multidimensional, involving various atmospheric, oceanic, and land variables collected over extensive spatial and temporal scales. Optical flow, primarily designed to capture motion between frames with high-temporal granularity (usually in fractions of a second), struggles to adequately capture the intricate interactions and patterns present in geospatial data. Another critical factor that renders optical flow unsuitable for geospatial data analysis is the presence of deformable objects. Unlike traditional image sequences where objects have consistent pixel-level motion, geospatial data involves dynamic elements such as clouds, atmospheric fronts, and ocean currents, which can undergo complex and non-rigid deformations. Optical flow-based methods assume pixel motion arises from rigid body movements, making them ill-suited for interpolating geospatial data. As a result, leveraging optical flow for geospatial data interpolation might lead to oversights and inaccuracies, emphasizing the need for specialized methods to effectively address the unique characteristics and challenges presented by geospatial datasets.

To summarize, temporal interpolation between successive frames in video sequences is a critical component with diverse applications. While supervised and unsupervised methods for temporal interpolation have shown some promising outcomes for video data, the state-of-the-art still relies on an optical flow formulation encoding the motion field of pixels. The low temporal granularity paired with dynamic spatial qualities, with a range of movements, deformations, and temporal alterations of geospatial datasets defy several assumptions inherent to optical flow. In response to these challenges, we introduce STint (Self-supervised Temporal Interpolation), a temporal interpolation approach that avoids the need for optical flow. STint leverages a unique dual cycle consistency training mechanism that is built upon the idea of cycle consistency from Figure~\ref{fig:cc}, eliminating the need for labels or dense motion data while efficiently harnessing spatiotemporal context for interpolation tasks. To summarize, our contributions are as follows:
\begin{itemize}
    \item We propose STint- a self-supervised framework that exploits a reformed dual cycle consistency constraint for temporal interpolation.
    \item STint learns to temporally interpolate without using any motion information like optical flow, thus making it agnostic to any motion/occlusion-related assumptions.
    \item We present the effectiveness of STint's self-supervised dual cycle consistent training paired with its optical flow-agnostic property on various geospatial datasets. 
\end{itemize}

\section{Related Work}
In this section, we discuss the topics relevant to this work, starting from methods for video frame interpolation and comparing our proposed method with them. Next, we discuss a few methods applying the cycle consistency constraint (shown in Figure~\ref{fig:cc}) that inspired us to develop our proposed dual cycle consistency technique for STint.

\subsection{Video Frame Interpolation}
In recent years, significant advancements have been made in the field of video interpolation. The SuperSloMo method by Jiang et al.~\cite{superslomo}  is a supervised approach that uses a two-stage process involving optical flow estimation and adaptive convolution layers for synthesizing intermediate frames. Niklaus et al.~\cite{niklaus2020softmax} proposed a deep learning framework for video frame interpolation using asymmetric reverse flow and cyclic fine-tuning, which models bidirectional motion and refines generated frames iteratively. Despite the methods being new and offering improved performance, they still rely on optical flow and may face generalization issues in scenarios discussed in previous sections. Liu et al.~\cite{dvf} developed DVF, a deep learning-based video frame interpolation network that exploits flowing pixel values (voxel) from a given frame to the target frame. While the idea of using an augmented version of optical flow seems better, it is bound by the assumption that most of the pixel patches between frames are near copies of each other, which may not be applicable for data types other than high-frame-rate videos. There exists other methods~\cite{jin2023enhanced, park2021asymmetric, danier2022st} that follow similar ideas and incorporate motion estimation for frame synthesis. However, like other optical flow-based methods, these will most likely fail for data types with non-rigid motion and/or lower temporal granularity, both of which are generally the characteristics of geospatial data. The idea of optical flow agnostic video interpolation has been investigated recently. Specifically, FLAVR~\cite{flavr} was proposed as a flow-agnostic method that is end-to-end trainable for multi-frame video interpolation. However, FLAVR~\cite{flavr} is limited to domains where direct supervision is available. Our work presents an unsupervised framework for temporal interpolation without the need for optical flow, thus making it specifically applicable to geospatial data types.

\subsection{Unsupervised Cycle Consistency}
When we talk about unsupervised cycle consistency, we are addressing a property that assures when we transition from one domain to another, and then by reverting to the original, we end up close to our starting point. For instance, in a task such as translating an image from a source to target style~\cite{zhu2017unpaired} (e.g., transforming a daytime image to appear as if it was taken at night) and then translating it back to the original style, the cycle consistency loss measures the dissimilarity between the original and the twice-translated image. The concept of unsupervised cycle consistency has shown some promising potential in video interpolation tasks~\cite{deepcycle, unsupcycle}.  In the context of video interpolation, cycle consistency can be seen as trying to reconstruct back frames from the original temporal resolution by interpolating between predicted intermediate frames. Liu et al.~\cite{deepcycle} was the first to introduce cycle consistency as a constraint to regularize a fully supervised video interpolation model. Most related to our wor is~\cite{unsupcycle}, which used cycle consistency for unsupervised pre-training on SuperSloMo~\cite{slomo} architecture for video interpolation. Our work notably differs in several essential respects. The most important being that our technique does not require any dense motion information while being fully unsupervised to perform temporal interpolation, as compared to ~\cite{deepcycle, unsupcycle}, both of which rely on estimating optical flow to model dense pixel movements. Secondly, our technique employs an improved dual cycle consistency loss which not only attempts to reconstruct the original sequence but also regularizes the predicted intermediate sequence at appropriate time stamps for better overall interpolation. 

\section{Method}
By introducing STint, we aim to contribute a novel approach that operates under two key rules - unsupervised learning and motion-agnostic enhancement. The subsequent subsections delve into the methodology and technical details of the proposed STint.

\begin{figure*}[t]
\centering
\includegraphics[width=0.95\textwidth]{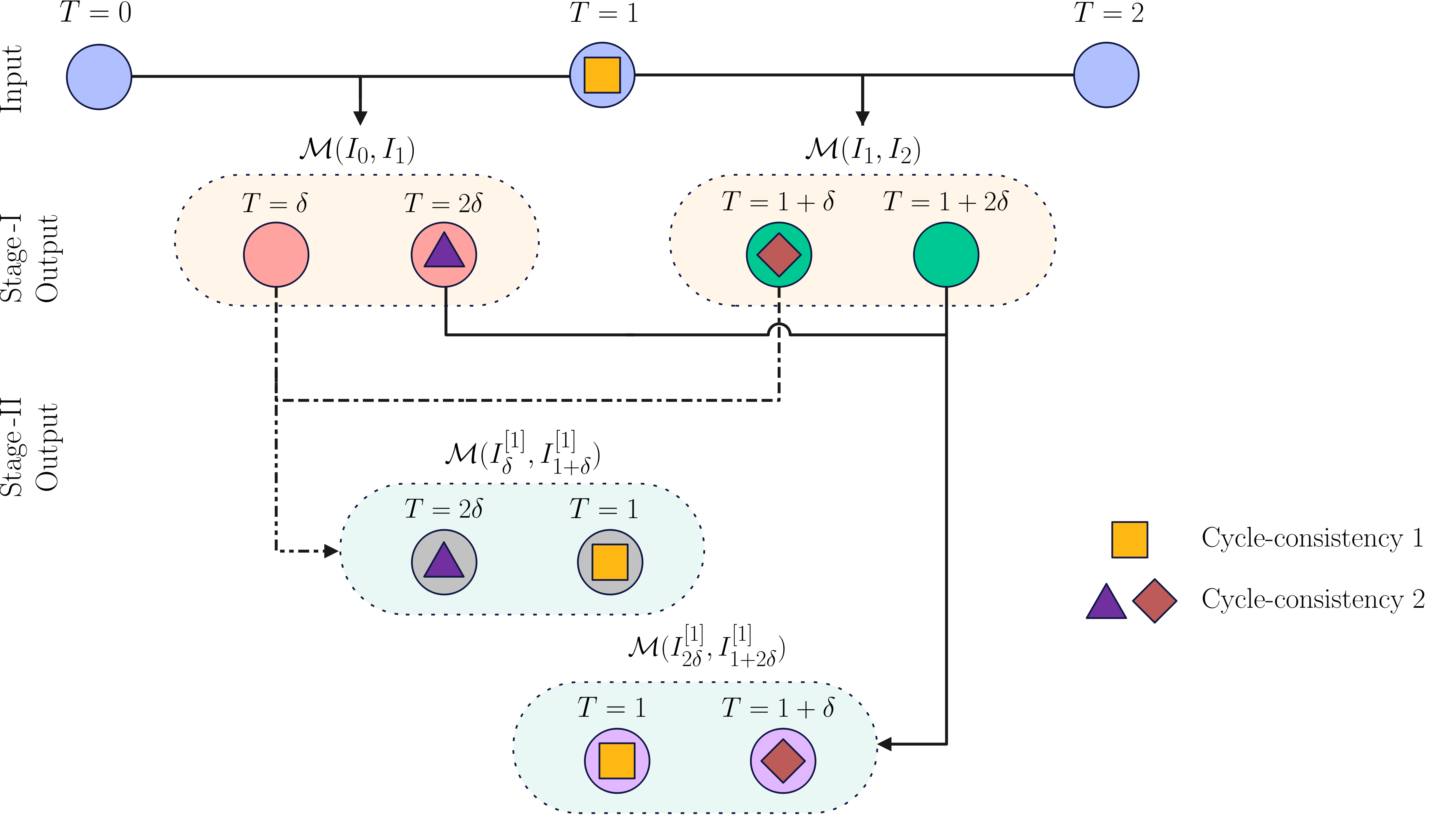} 
\caption{A detailed overview of the STint framework. Building upon the cycle consistency pretext task from Figure~\ref{fig:cc}, we propose to learn our model $\mathcal{M}$ for temporal interpolation via multiple cycle consistency losses (indicated in inscribed shapes), obtained by interpolating two frames between consecutive input frames across multiple stages. Our proposed dual cycle consistency makes all the transformations predicted by the model to be invertible.}
\label{fig:cc_main}
\end{figure*}

\subsection{Dual Cycle Consistent Interpolation}
We consider the following learning setting: We are given spatiotemporal data $D_K$ with $K$ as the temporal frequency. We are interested in increasing this temporal frequency using our proposed STint with the following properties: (a) It is fully unsupervised, and (b) It does not use motion information, i.e., optical flow. We consider the task of generating two additional frames between each pair in the original sequence. Specifically, consider  $[I_0, I_{1}, I_{2}]$ to be a triplet of consecutive input frames from a spatiotemporal sequence $\{I_{T_i}\}^N_{i=0}$, we aim to learn a network that takes two consecutive frames as inputs and can predict two intermediate frames in between these consecutive input frames. Considering $[I_0, I_{1}]$ as inputs, we obtain the corresponding intermediate sequence $[I_{\delta}, I_{2\delta}]$ as follows:   
\begin{equation}
 [I_{\delta}, I_{2\delta}] = \mathcal{M}(I_0, I_1)
\label{eq:1}
\end{equation}
where $\mathcal{M}$ denotes the network that we aim to train for temporal interpolation. Our goal is to train $\mathcal{M}$ in an unsupervised manner, that is, without relying on pre-labeled training data or explicit guidance during the learning process. 

Motivated by the recent strides made in video interpolation~\cite{deepcycle, unsupcycle}, as well as the remarkable progress in unpaired image-to-image translation using Generative Adversarial Networks (GANs)~\cite{zhu2017unpaired}, we propose to use a similar approach for refining our network $\mathcal{M}$ for temporal interpolation. The cornerstone of our approach is the preservation of dual cycle consistency over the duration of the entire temporal sequence. Cycle consistency, in essence, means that when we execute an operation and its inverse consecutively, we return to our original starting position. For instance, in the context of our network $\mathcal{M}$ for temporal interpolation, we aim for a scenario where a series of transformations applied to the sequence frames, followed by their reverse transformations, should produce a result that closely resembles the initial temporal sequence. We achieve this via multiple cycle consistency losses across different stages:

\subsubsection{Stage I:} Our unsupervised dual cycle consistency constraint consists of two phases, as shown in Figure~\ref{fig:cc_main}. The stage-I begins by considering two pairs of consecutive input frames $[I_0, I_1]$ and $[I_1, I_2]$; therefore, the individual corresponding intermediate frames using a network $\mathcal{M}$ can be written as:
\begin{equation}
       [I_{\delta}^{[1]}, I_{2\delta}^{[1]}] = \mathcal{M}(I_0, I_1) 
\end{equation}
\begin{equation}
       [I_{1+\delta}^{[1]}, I_{1+2\delta}^{[1]}] = \mathcal{M}(I_1, I_2) 
\end{equation}
\subsubsection{Stage II:} The idea here is to use the predicted intermediate frames from stage I with the temporal window that matches the original sequence as input pairs again. In this particular case, the idea is to use $I_{\delta}^{[1]}, I_{1+\delta}^{[1]}]$ and $[I_{2\delta}^{[1]}, I_{1+2t}^{[1]}]$ as inputs. Mathematically, it can be written as:
\begin{equation}
   [I_{2\delta}^{[2]}, I_{1}^{[2]}] = \mathcal{M}(I_{\delta}^{[1]}, I_{1+\delta}^{[1]})
\end{equation}
\begin{equation}
   [I_{1}^{[2]}, I_{1+\delta}^{[2]}] = \mathcal{M}(I_{2\delta}^{[1]}, I_{1+2\delta}^{[1]})
\end{equation}

The first trick is that both the predictions at $T=1$ from stage II must match the original middle input frame $I_1$, thus we introduce the first cycle consistency constraint to optimizing the network $\mathcal{M}$  on the reconstruction error at $T=1$ as follows:
\begin{equation}
\mathcal{L}_{CC_1} = arg\, min_{\mathcal{M}_\theta} (|| \,I_1 - \mathcal{M}(I^{[1]}_\delta, I^{[1]}_{1+\delta})(2)\,||_1 + ||\,I_1 - \mathcal{M}(I^{[1]}_{2\delta}, I^{[1]}_{1+2\delta})(1)\,||_1)
\end{equation}
where $\mathcal{M}(\cdot)(p), p\in[1,2]$ refers to $p^{th}$ prediction out of two outputs from $\mathcal{M}$. $\mathcal{L}_{CC_1}$ corresponds to the yellow squares shown in Figure~\ref{fig:cc_main}.

To make the interpolation more stable, we propose to integrate cycle consistency constraint on the other predictions from stage II. Specifically, we enforce similarity between predictions from stage II (${I}^{[2]}_{2\delta}$ \& ${I}^{[2]}_{1+\delta}$) and predictions from stage I ($I^{[1]}_{2\delta}$ \& $I^{[1]}_{1+\delta}$), shown as inscribed triangle and diamond in Figure~\ref{fig:cc_main}. Mathematically, we can write it as:

\begin{equation}
\begin{split}
    \mathcal{L}_{CC_2} &= arg\, min_{\mathcal{M}_\theta} \frac{1}{2}\left(||\, I_{2\delta}^{[1]} - {I}_{2\delta}^{[2]}\,||_1 + ||\,I_{1+\delta}^{[1]} - I_{1+\delta}^{[2]} \,||_1\right) \\
    &= arg\, min_{\mathcal{M}_\theta} \frac{1}{2}(||\mathcal{M}(I_1, I_2)(2) - \mathcal{M}(I^{[1]}_{\delta}, I^{[1]}_{1+\delta})(1)||_1 \\&\;\;\;\;\;\;\;\;\;\;+ ||\,\mathcal{M}(I_1, I_2)(1) - \mathcal{M}(I_{2\delta}, I_{1+2\delta})(2)\,||_1 )
\end{split}
\end{equation}

While the first cycle consistency constraint ($\mathcal{L}_{CC_1}$) acts as the key for learning to interpolate between frames, the second cycle consistency term ($\mathcal{L}_{CC_2}$) would help generate stable intermediate frames such that it maintains both spatial and temporal consistency with the entirety of the sequence. This dual cycle consistency procedure can be applied to all such intermediate frames, making the learning procedure complete. Optimizing on this dual cycle consistency constraints the model to learn to produce plausible intermediate climate frames so that these can then be used to reversely reconstruct the original input frames.

\begin{equation}
\mathcal{L} = \lambda_{CC_1} \mathcal{L}_{CC_1} + \lambda_{CC_2} \mathcal{L}_{CC_2}      
\end{equation}

\textbf{Avoiding Trivial Solutions:} A straightforward solution to the dual cycle consistency method as defined above could be to merely duplicate the input frames for intermediate predictions. However, this does not transpire in our learning setting because we repeatedly compel the model to create reversible frames for pairs comprising diverse temporal frames. Moreover, no path exists for transmitting the temporal information of the input frames between successive forward propagations, thus making the learning avoid trivial solutions.

\subsection{Interpolation Network \texorpdfstring{$\mathcal{M}$}{M}}
\begin{figure}
    \centering
    \includegraphics[width=0.65\linewidth]{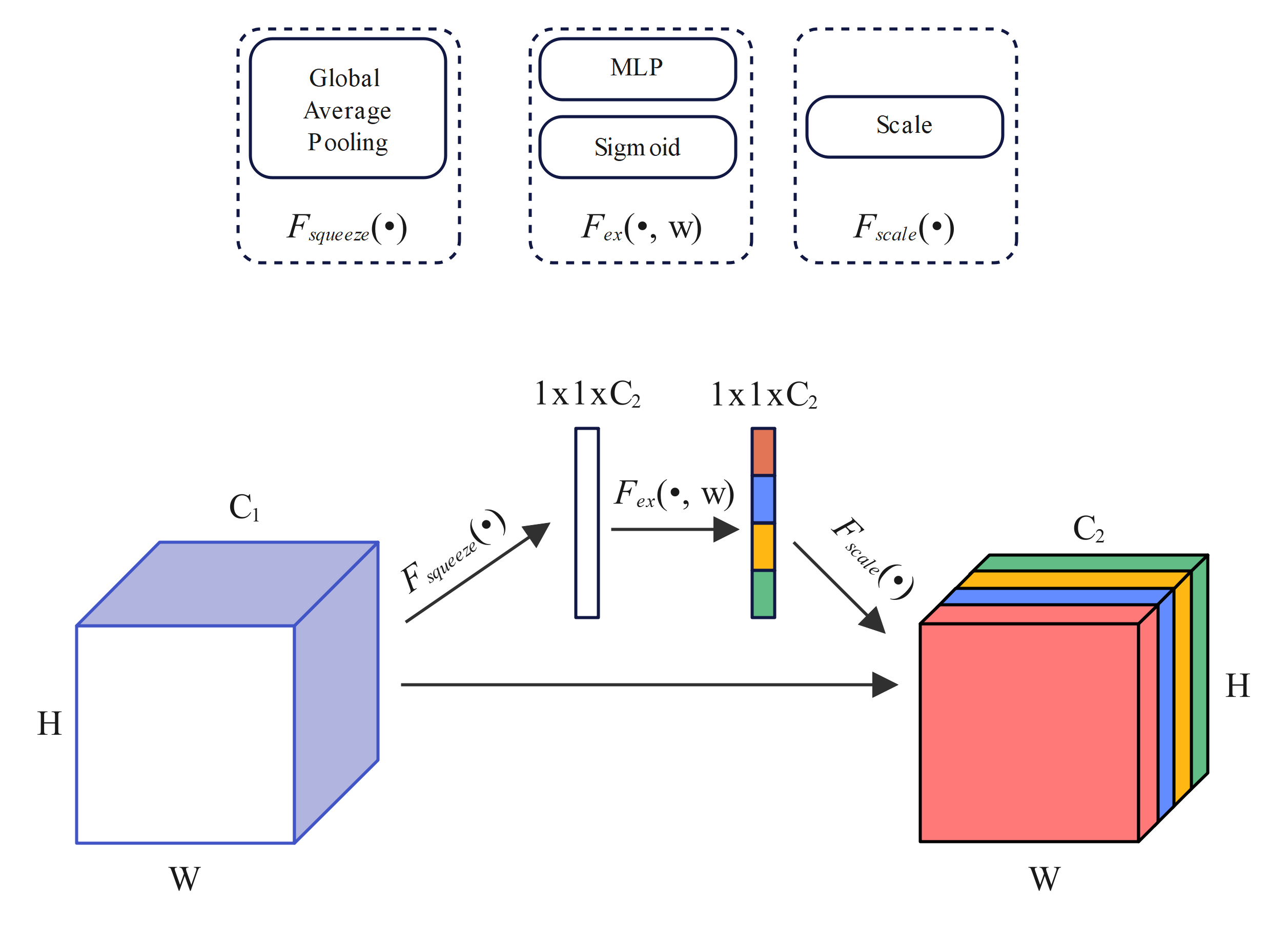}
    \caption{Squeeze and Excitation Block}
    \label{fig:se}
\end{figure}

Several existing methods, such as ConvLSTMs~\cite{shi2015convolutional}, 3D ResNet~\cite{tran2018closer}, PyConv~\cite{duta2020pyramidal} and SepConv~\cite{niklaus2017video}, can serve as the baseline model. In this work, we choose a modified version of 3D-U-Net~\cite{cciccek20163d} due to its notable simplicity, paired with its impressive learning capabilities. 3D-U-Net~\cite{cciccek20163d} is designed by adapting the original U-Net~\cite{ronneberger2015u} stem to 3D inputs by replacing the 2D convolutions with consecutive 3D convolutional layers. We go a step ahead and adopt the following lightweight architectural changes to improve model performance.
\begin{itemize}  
    \item We remove the max-pooling layers from the original 3D-U-Net while retaining the 3D convolution and transposed convolution layers.
    \item We incorporate Squeeze-and-Excite (SE) blocks~\cite{hu2018squeeze} shown in Figure~\ref{fig:se} into 3D-U-Net. We adapt these SE blocks to spatiotemporal inputs by simply using a 3D global average pooling operation for the squeeze function, making them 3D SE blocks. 
\end{itemize}
Henceforth, with these modifications, we consider our network $\mathcal{M}$ for temporal interpolation as 3D-U-Net-SE. The primary driver for integrating 3D SE blocks into our model was to enhance the quality of representations generated by the original 3D-U-Net. These 3D SE blocks contribute to better modeling of inter-dependencies among the convolutional features and carry out feature re-calibration~\cite{hu2018squeeze}. Through this re-calibration, the model can leverage global information to selectively accentuate informative features while diminishing the less useful ones, thereby improving overall performance.

\subsection{Unsupervised Training and Fine-tuning}
For any given data, we first obtain triplets of frames from a given sequence such that two intermediate frames exist between every consecutive frame. Each consecutive pair from this triplet is then used as an input to the STint framework with a focus on maintaining dual cycle consistency as described in the previous section.  Unless mentioned otherwise, the loss coefficients $\lambda_{CC_1}$ and $\lambda_{CC_2}$ have been set empirically using a validation set as $\lambda_{CC_1} = 0.65$, $\lambda_{CC_2} = 0.35$, 

Following the initial unsupervised training, we fine-tune our mode by using the model's learned parameters as a starting point and further optimizing these parameters. This process involves a few additional rounds of training with a lower learning rate to refine the parameters without dramatically altering the existing structure of the learned features. This ensures that the model progressively learns to capture more intricate details and nuances, thereby boosting the quality of the interpolated frames. For fine-tuning, we use an optimization criterion ($\mathcal{L}_{fine}$) that is a weighted combination of the reconstructed errors from ground truth and the dual cycle consistency losses. Mathematically, it can be written as:
\begin{equation}
    \mathcal{L}_{fine} = \mathcal{L}_{||I_t -\hat{I}_t ||} + \gamma_{CC_1} \mathcal{L}_{CC_1} + \gamma_{CC_2} \mathcal{L}_{CC_2}
\end{equation}
where $I_t$ \& $\hat{I}_t$ corresponds to the ground truth intermediate frames and the predictions respectively.

We found that including the dual cycle consistency loss along with the supervised reconstruction loss leads to faster and better convergence. While the supervised reconstruction loss $\mathcal{L}_{||I -\hat{I}_t ||}$ helps the model get direct feedback on the produced interpolated frames, the dual cycle consistency loss $\mathcal{L}_{CC_1}, \mathcal{L}_{CC_2}$ acts as a constraint that measures the difference between an original frame and a frame that has been interpolated and then reverse-interpolated. Minimizing this type of loss helps the model learn to produce interpolated frames that, when reverse-interpolated, closely resemble the original frames, thus making the model generate temporally consistent frames. Finally, we use experimentally selected weights for this linear combination of losses: $\gamma_{CC_1}=0.5$, $\gamma_{CC_2}=0.3$.

\section{Experiments}
In this section, we report the datasets used in the experiments and comparisons with the baseline and state-of-the-art. We also discuss the limitations of the proposed method. In all the following experiments, we report the test error from the epoch with the lowest validation error. 

Following previous works, we use PSNR and SSIM metrics to report the quantitative results of our method. For some geospatial datasets, we also report a criterion called Scatter Index (SI), also denoted as $\frac{MSE}{Capacity}$ which can be considered as a normalized version of \textit{Mean Square Error} (MSE), where \textit{Capacity} stands for the maximum value of the data variable under consideration~\cite{willmott1985statistics}.

\subsection{Architecture Details}
Built entirely on PyTorch~\cite{Pytorch}, the architecture of STint consists of approximately 40M parameters. Our backbone is a modified version of 3D-U-Net~\cite{cciccek20163d} by incorporating Squeeze-and-Excite (SE) blocks~\cite{hu2018squeeze} into 3D-U-Net~\cite{cciccek20163d}. Our choice of architecure is based on the simplicity of 3D-U-Net and the powerful learning capabilities offered by SE. Additionally, we find that replacing the max-pooling layers with 3D global average pooling in 3D-U-Net version by ~\cite{cciccek20163d} performs best for our experiments. We use ReLU between 3D convolutions and  3D transposed convolutions. Following~\cite{cciccek20163d}, we also introduce batch normalization (``BN'') before each ReLU. Unless specified otherwise, we use Adam for optimization with $\beta_1 = 0.9$, $\beta_2 = 0.999$, and no weight decay.

\subsection{Baseline \& State-of-the-Art} 
We establish our baseline as a straightforward approach of `trivial copy'. It refers to duplicating existing frames within the video sequence to generate interpolated frames. It mimics the temporal progression by copying frames at regular intervals without any complex computations or algorithms. For example, each sequence element of geospatial data at 3-hourly intervals can be replicated twice in a consecutive manner to obtain 1-hourly data.

Furthermore, considering that there are not a lot of unsupervised methods for the task of temporal interpolation except~\cite{unsupcycle}, which uses dense optical flow for frame interpolation. We choose~\cite{unsupcycle} for comparison. In scenarios where the optical flow was futile (more in subsequent subsections), some hyperparameters like learning rate needed to be tweaked in order to make the training possible; however, mostly all the settings were chosen to match the setup of~\cite{unsupcycle}.

\begin{table*}
\centering
\begin{tabular}{ccccc}
\toprule
Data source & Data type & Total frames & Spatial resolution & Temporal resolution\\
\midrule 
IPSL & Surface wind & 186,880  &  $96\times 236$ & 3-hourly\\
ERA5 & Solar-irradiance  & 196,375   & $96\times 236$ & 1-hourly\\
CARRA & Surface temperature & 190,224  & $200 \times 500$ & 1-hourly\\
SEN12MS & Cloud (Satellite) & 60,117 & $256\times256$   & 2-hourly \\
UCF-101 &  Human action videos & 120,790 & $256\times256$ & 25 FPS   \\
\bottomrule
\end{tabular}
    \caption{Statistics of datasets used for experiments.}
    \label{tab:data_stats}
\end{table*}

\subsection{Climate Geospatial Data}
Our experiments on geospatial domains encompass different atmospheric variables. Specifically, we consider three of the most commonly used sources for atmospheric geospatial data: 
\begin{itemize}
    \item \textbf{IPSL:} We use the 3-hourly resolution climate model outputs of the wind field from Institut Pierre-Simon Laplace (IPSL) Phase 6 of the Coupled Model Intercomparison Project (CM6A). IPSL is considered to have high fidelity in reproducing 20th-century observed climate over the US~\cite{boucher2020presentation}.
    \item \textbf{ERA5:}  We use the 1-hourly downward solar irradiance fields from ERA5~\cite{hersbach2020era5}, which is the fifth generation high-resolution observational data (reanalysis) from European Centre for Medium-Range Weather Forecasts (ECMWF). The ERA5 global reanalysis is the most used data set from Copernicus Climate Change Service (C3S) in the Copernicus Climate Data Store (CDS)~\cite{hersbach2020era5}.
    \item \textbf{CARRA:} We consider 1-hourly regional observational records of surface temperature for the Arctic region   ~\cite{koltzow2022value}. While the ERA5 global reanalysis serves as a widely used dataset, its broad scope does not cater to the intricate nuances of the Arctic's rapidly changing climate. This makes CARRA an ideal candidate to study temporal interpolation in this geographically challenging subcontinent.
\end{itemize} 
For each quadruple of consecutive frames from the spatiotemporal sequence, the middle two frames serve as the ground truth while the other two are inputs. Table~\ref{tab:data_stats} summarizes the statistics of each dataset. We use $80$-percent of the total frames for unsupervised training and fine-tuning, while the rest is used for evaluation.

\subsubsection{STint Pre-training \& Finetuning}
We use a combination of four NVIDIA Quadro RTX 5000 and an RTX 3090 to train the model for $400$ epochs with a learning rate with a batch size of 16. We initialize the learning rate with $3\times 10^{-4}$ and decay it by a factor of $2$ for every $400,000$ mini-batch updates or when the performance stagnates. We use Adam optimizer with coefficients $(0.9, 0.999)$ with a weight decay of $0.05$ to optimize the loss. We find that the cycle-consistency loss weight values of $\lambda_{CC_1} = 0.65$, $\lambda_{CC_2} = 0.35$ work best for our setting and use it throughout the experiments. We load the pre-trained weights and fine-tune STint in an end-to-end manner. We train the model for 50 epochs with a learning rate of $2\times 10^{-3}$. We perform optimization on a linear combination of losses cycle consistency losses with the error from supervision. We use experimentally selected weights for this linear combination of losses, where the weights for cycle consistency loss are: $\gamma_{CC_1}=0.5$, $\gamma_{CC_2}=0.3$.

\begin{table}
\centering
\begin{tabular}{cccc}
\toprule
\multicolumn{4}{c}{\textbf{ERA5 Solar}} \\
\midrule
    & $\frac{MSE}{Capacity}$ ($\downarrow$)& PSNR ($\uparrow$) & SSIM ($\uparrow$)\\
\midrule
Baseline   &  0.3086  & 25.238 & 0.623     \\
SuperSloMo      &  0.0907        & 30.157       &    0.733\\
Proposed       &  \textbf{0.0561}  &  \textbf{32.731} & \textbf{0.792} \\
\midrule
\multicolumn{4}{c}{\textbf{IPSL Wind}} \\
\midrule
Baseline   &  0.6206  &  24.097  & 0.619\\
SuperSloMo       &    0.4150&  29.713  & 0.681 \\
Proposed       &  \textbf{0.2904}& \textbf{31.167} & \textbf{0.713}\\
\midrule

\multicolumn{4}{c}{\textbf{CARRA Temperature}} \\
\midrule
Baseline   &  0.5319   &  27.832 & 0.667\\
SuperSloMo       &  0.1604   &  30.276  &0.724\\
Proposed       &  \textbf{0.0975}& \textbf{31.908}   &\textbf{0.775}\\
\bottomrule
\end{tabular}
\caption{Temporal interpolation results on geospatial data with different atmospheric variables.}
\label{tab:climate}
\end{table}

\begin{figure}
    \centering
    \includegraphics[width=0.7\linewidth]{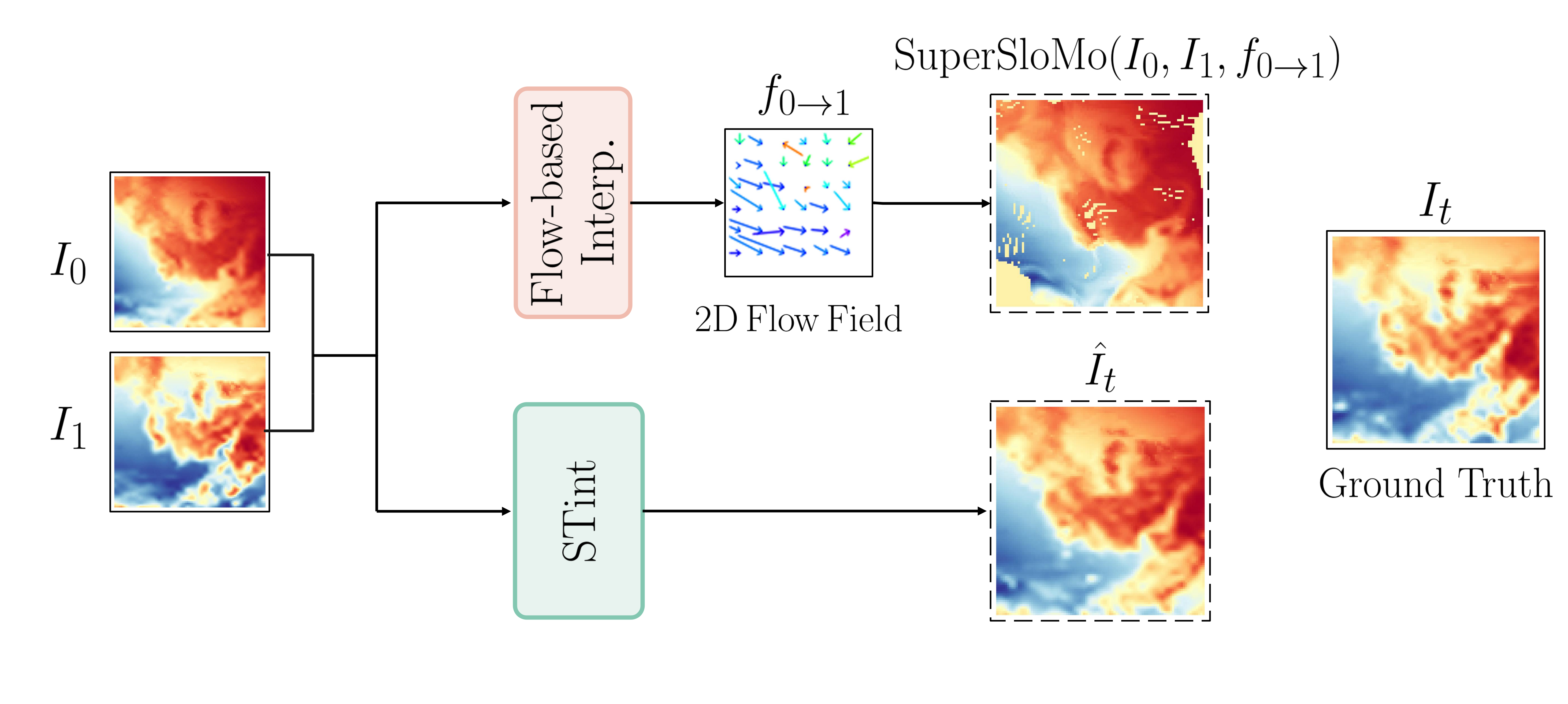}
    \caption{An example of temporal interpolation of ERA5 solar field data. Optical flow-based SuperSloMo~\cite{unsupcycle} can be seen to struggle with the non-rigidity of motion inherent to solar data. Our proposed STint, a fully unsupervised and optical flow agnostic technique, generates intermediate frame $\hat{I}_t$ that is much closer to the ground truth $I_t$, making it a robust interpolation method in scenarios with non-rigid motion.}
    \label{fig:intro}
\end{figure}

Table~\ref{tab:climate} presents our results which show that our unsupervised dual cycle consistency-based STint outperforms SuperSloMo~\cite{unsupcycle} uses regular cycle consistency (described in Figure~\ref{fig:cc}) for self-supervised training. We observe consistent performance gains across all the climate variables. In contrast to cycle consistency-based methods, predictions using baseline trivial copy result in notably inferior interpolation frames. Using IPSL wind fields data, at a frame count three times larger than both ERA5 and CARRA datasets, we outperform SuperSloMo~\cite{unsupcycle} with PSNR of 31.17dB vs. 29.71dB. Figure~\ref{fig:intro}
shows a visual comparison of the predictions on ERA5 solar field data. The highly dynamic temporal nature, followed by extensive deformations of spatial features between frames, falters the optical flow estimation, thus generating artifacts while reconstructing the intermediate frames. STint, on the other hand, generates predictions that are much closer to the actual ground truth intermediate frames.

\begin{figure*}[t]
\centering
\includegraphics[width=0.95\textwidth]{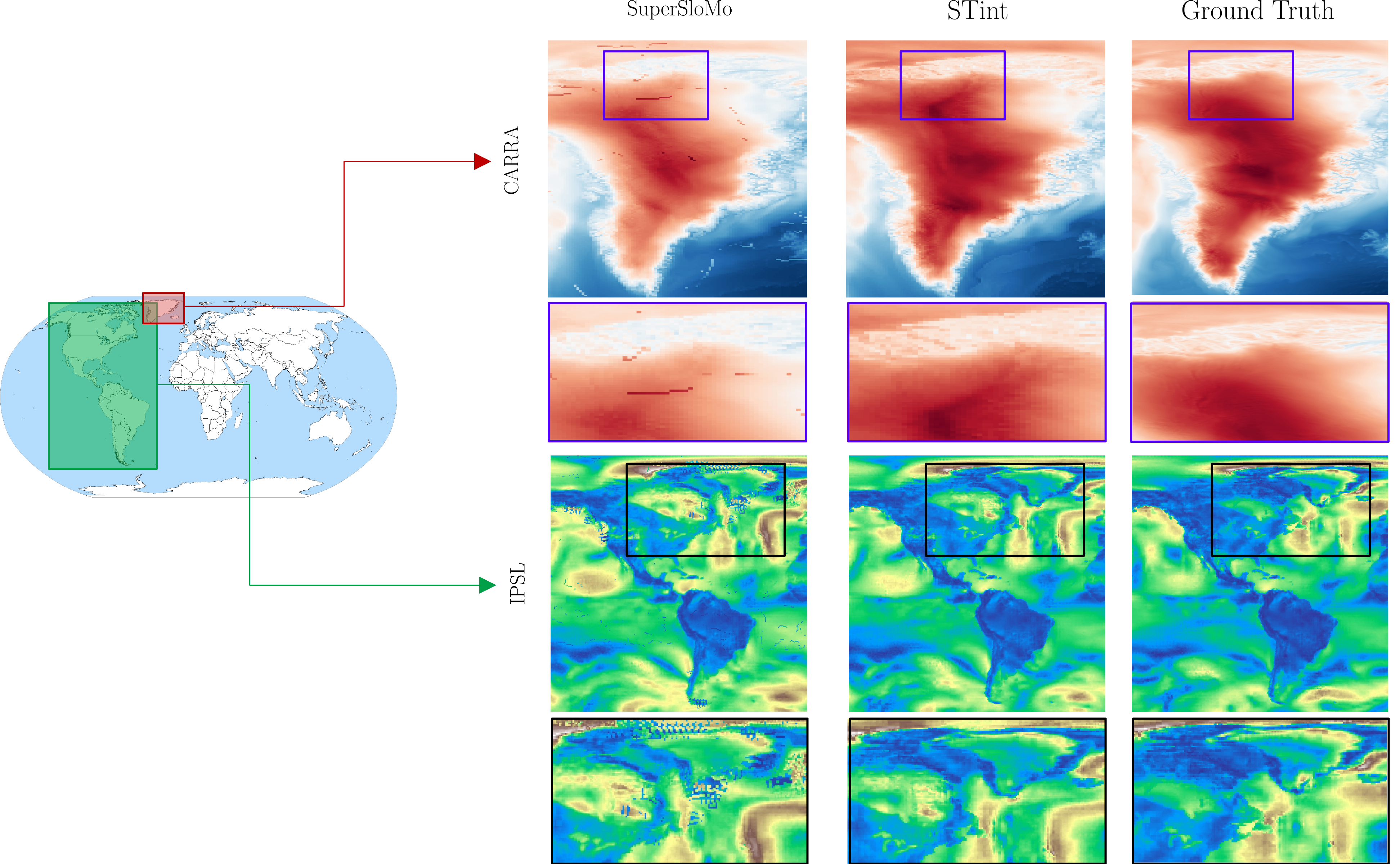} 
\caption{Qualitative comparison of a sample of predicted intermediate frames from STint vs SuperSloMo~\cite{unsupcycle} on IPSL~\cite{boucher2020presentation} and CARRA~\cite{koltzow2022value} geospatial datasets. The regions with dynamic motions lead to tearing artifacts in optical flow-based SuperSloMo~\cite{unsupcycle} but are predicted well with STint.}
\label{fig:supp_vis}
\end{figure*}

Since the quantitative measures show only a weak correlation with the human visual system, Figure~\ref{fig:supp_vis} visually compares the predicted intermediate frames on IPSL~\cite{boucher2020presentation} and CARRA~\cite{koltzow2022value} from different methods. Optical flow-based SuperSloMo struggles at regions with  high deformations, complex and large movements~\cite{brox2004high}. This can be seen in the form of tearing artifacts in the predicted intermediate frames. The predictions from STint are visually much closer to the ground truth.   

\subsection{Domain Transfer to Satellite Imagery}
We further study the effectiveness of our proposed dual-cycle consistency for pre-training and reducing gaps for domain transfer. For this study, we consider self-supervised pre-training on ERA5 solar fields followed by fine-tuning on SEN12MS cloud imagery dataset~\cite{schmitt2019sen12ms}. While solar radiation and cloud imagery may seem like two distinct domains without a connection, cloud cover acts as spatial filters impacting solar radiation by altering its spatial properties~\cite{tzoumanikas2016effect}. Table~\ref{tab:satellite} shows the results indicating the pre-trained model fine-tuned on SEN12MS outperforms the one trained only on SEN12MS in a supervised way with random initialization. This can be attributed to the vulnerability of supervised methods to performance drops in data-scarce scenarios. Conversely, unsupervised pre-trained models, having learned from data-rich scenarios, exhibit adaptability through domain transfer, proving advantageous in data-poor regions such as with SEN12MS data.
While unsupervised pre-training improves performance for both SuperSloMo~\cite{unsupcycle} and STint, our proposed STint still outperforms the former, demonstrating that our method's optical-flow agnostic property generalizes well to satellite imagery. 

\subsubsection{Domain Adaptation to SEN12MS}
In this section, we examine a data-poor scenario of another geospatial data, that is SEN12MS cloud Imagery (60K frames),  where labeled data for interpolation is very less for training. However, an abundance of data is available in another domain, that is ERA5 (196K frames). Our model evaluation takes place on the SEN12MS dataset. For each set of four frames, we utilize the first and fourth frames as input to predict the second and third frames. We use the same model as above, but we utilize the model pre-trained on ERA5 to fine-tune it on the SEN12MS dataset for 50 epochs.  We use a similar learning rate and weight decay factor for fine-tuning as described in the previous experiment. For the model trained only with supervision, we use 3D-U-Net-SE with random initialization and train it for 200 epochs with a learning rate of $2\times 10^{-3}$.

\begin{table}
\centering
\begin{tabular}{ccc}
\toprule
\multicolumn{1}{c}{} & \multicolumn{1}{c}{\textbf{ERA5 $\rightarrow$ SEN12MS}} & \multicolumn{1}{c}{\textbf{SEN12MS}} \\
\midrule
    &  PSNR/SSIM & PSNR/SSIM \\
\midrule
\centering
Baseline    &   20.66 /0.513 &  20.66 /0.513  \\
SuperSloMo        &   26.71 /0.612  & 24.83 /0.559  \\
Proposed        &   \textbf{28.18 /0.647}   &    \textbf{26.97 /0.594}   \\
\bottomrule
\end{tabular}
\caption{Results of Domain transfer results comparing domain transfer performance on SEN12MS cloud imagery dataset against model trained in a supervised manner.}
\label{tab:satellite}
\end{table}

\subsection{Limitation: Video Data}
\begin{wraptable}{r}{0.4\textwidth}
\begin{tabular}{ccc}
\toprule
\multicolumn{3}{c}{\textbf{UCF101}} \\
\midrule
    &  PSNR ($\uparrow$) & SSIM ($\uparrow$)\\
\midrule
Baseline        &   30.14 &  0.877\\
SuperSloMo           &   \textbf{34.19} & \textbf{0.917}  \\
Proposed            &    32.52  &  0.882 \\
\bottomrule
\end{tabular}
\caption{Temporal interpolation results on UCF101 action recognition dataset with 25 frames-per-second.}
\label{tab:ucf}
\end{wraptable}
We test self-supervised training of our approach against SuperSloMo~\cite{unsupcycle} on UCF101 dataset~\cite{soomro2012dataset}. We follow the same procedure from previous experiments of temporal sub-sampling. For a set of four frames, The first and the fourth frames are used as input to predict the middle two frames. For each frame, we use standard normalization. For the purpose of data augmentation, following~\cite{unsupcycle}, we leverage the problem's symmetry by randomly choosing input sequences during training and reversing the temporal order of the frames. We pre-train the model from random initialization for 400 epochs with a batch-size of 16. We fine-tune the pre-trained model on the train set (75 percent) for 60 epochs and evaluate the performance on the disjoint test set (25 percent).  We report the average value of the metric over all the predicted frames. We use a combination of losses from supervision and cycle consistency, where the weights for cycle consistency are $\gamma_{CC_1}=0.4$, $\gamma_{CC_2}=0.25$.

The testing setting remains similar to the previous setting, that is, for every quadruple, the first and fourth frames are used to reconstruct the middle two frames. While the spatial dynamics of geospatial datasets might break the underlying assumptions of optical flow, the same pixel-level motion dynamics assists in generating visually plausible intermediate frames for video datasets as can be seen from results in Table~\ref{tab:ucf}.  Optical flow-based SuperSloMo~\cite{unsupcycle} achieves a gain in PSNR value of 1.67dB over STint. This implies that the effectiveness of optical flow-based SuperSloMo~\cite{unsupcycle} in benefiting temporal interpolation hinges on the validity of optical flow's assumptions, which holds for the UCF101 dataset with 25-frames-per-second but does not hold for geospatial datasets which usually exists at an hour-level-resolutions.   

\begin{figure}
    \centering
    \includegraphics{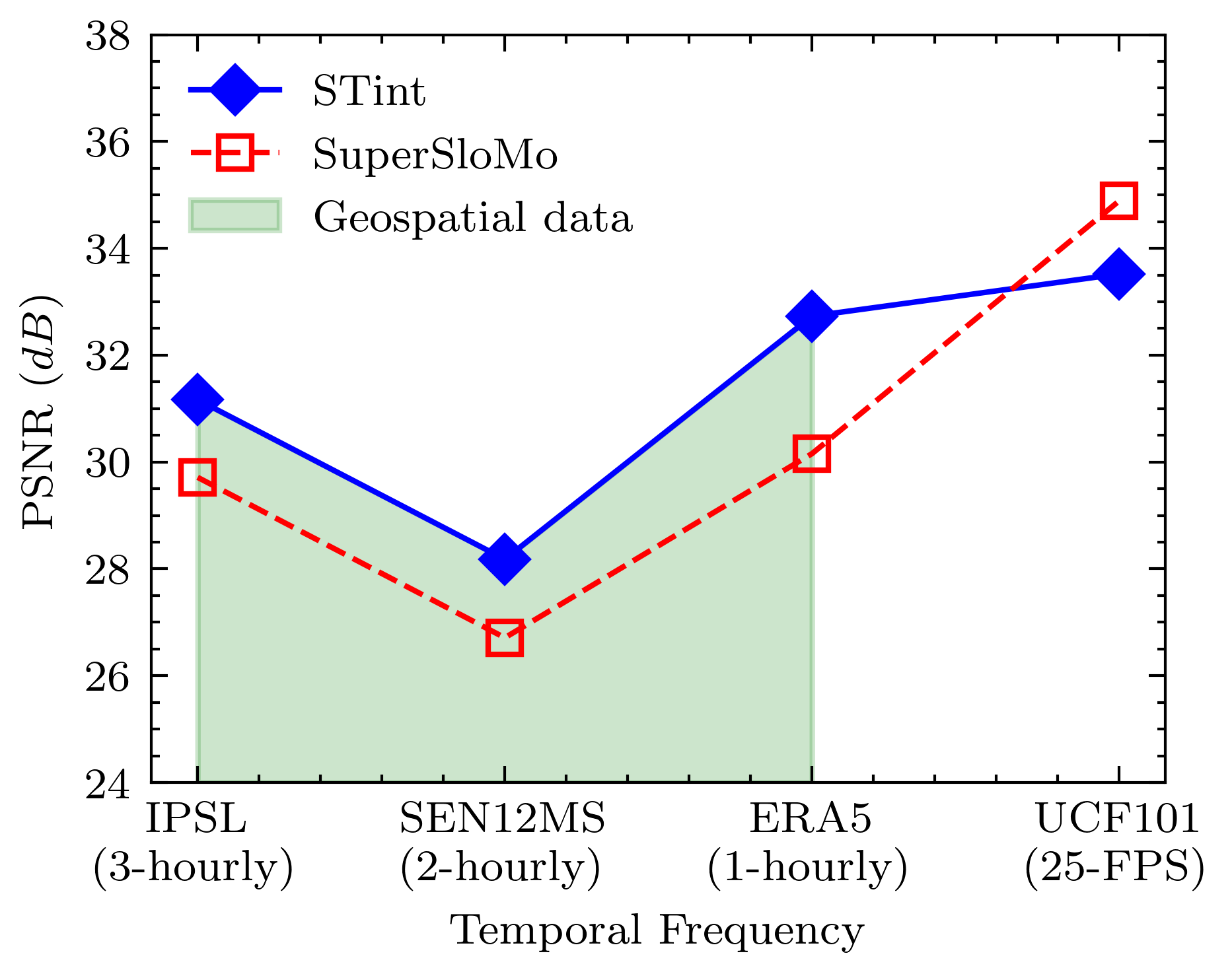}
    \caption{Performance comparison of STint vs SuperSloMo~\cite{unsupcycle} across datasets. Optical flow leads to better temporal interpolation for SuperSloMo~\cite{unsupcycle} on the video dataset of UCF101~\cite{soomro2012dataset}. The flow-agnostic property makes STint outperform SuperSoMo~\cite{unsupcycle} on all the geospatial datasets.}
    \label{fig:supp_psnr}
\end{figure}

Figure~\ref{fig:supp_psnr} compares the performance of STint and SuperSloMo across the datasets with different temporal frequencies. STint outperforms in scenarios with lower temporal granularity, while optical flow-based SuperSloMo benefits from the information of pixel movements for video data such as UCF101~\cite{soomro2012dataset}. 

\subsection{Physical Interpretation of cycle consistency}
Physical consistency and temporal evolution of geospatial data from our climate system are vital to understanding how our weather patterns and climate evolve over time. The climate system evolves over varying timescales, ranging from hourly to seasonal to centuries or even longer. The nonlinear dynamics of the system dictate a unique solution for the system's evolution over time, given various forcings and prior conditions. The physical consistency of these unique solutions to the system's equations ensures that the evolution of the system is consistent with known physical laws such as mass or energy conservation. Hence, given the evolution of the system over a chaotic attractor, the state of the system is over a unique trajectory, and hence, temporal interpolation of the system over this trajectory gives unique solutions on the attractor ~\cite{palmer2014real}. Ensuring that interpolation models are temporally coherent and consistent, such as STint is essential in improving our ability to predict changes in the geospatial systems over various timescales.

\section{Discussion}
This paper advances the state-of-the-art of temporal interpolation for geospatial data. The prevalent reliance on optical flow, while effective for video data, exhibits constraints when applied to geospatial data characterized by distinct temporal dynamics and movement patterns. In response, we have introduced a new approach termed Self-supervised Temporal Interpolation (STint). Unlike conventional techniques, STint does not rely on optical flow or ground truth data. Its novel dual cycle-consistency framework facilitates the generation of intermediate frames in a self-supervised manner. Through extensive experimentation across diverse geospatial datasets, we have established that STint offers a notable enhancement over existing methods.  By addressing the limitations of traditional approaches, STint contributes meaningfully to the domain of geospatial temporal interpolation. While our experiments have been tested by sampling from the original high-frequency dataset, STint can be simply employed to just temporally interpolate geospatial data in a self-supervised manner. 

Training using cycle consistency has its challenges, notably instability between batches. While training with a lower learning rate, in addition to certain regularizers, helps alleviate the instability during training, it leads to longer training periods. In the future, finding ways to make the cycle consistency technique more efficient would be useful. Additionally, since STint does not rely on optical flow, it helps in temporally interpolating geospatial datasets better, but there are certain scenarios like video data where motion information is more effective in interpolation. Moreover, it is also worth exploring the idea of an augmented version of optical flow that is applicable to geospatial and other scientific data. Lastly, we believe investigating better architectures for temporal interpolation will be key to generalizing dual cycle consistency for other data domains.

\section*{Acknowledgments}
This project was partially supported by the Climate Change AI Innovation Grants program, hosted by Climate Change AI with the support of the Quadrature Climate Foundation, Schmidt Futures, and the Canada Hub of Future Earth. This research was also partially funded through a grant provided by the National Science Foundation, under Award No. 2118285.
\bibliography{main}{}
\bibliographystyle{plain}
\medskip

\end{document}